\documentclass{article}

\usepackage[preprint]{neurips_2020}
\usepackage[utf8]{inputenc} 
\usepackage[T1]{fontenc}    
\usepackage{hyperref}       
\usepackage{url}            
\usepackage{booktabs}       
\usepackage{amsfonts}       
\usepackage{nicefrac}       
\usepackage{microtype}      
\usepackage{authblk}

\usepackage{float}
\usepackage{graphicx}       

\title{PCBDet: An Efficient Deep Neural Network Object Detection Architecture for Automatic PCB Component Detection on the Edge}

\begin{document}

\author[1*]{Brian Li}
\author[1*]{Steven Palayew}
\author[1]{Francis Li}
\author[1,2]{Saad Abbasi}
\author[2]{Saeejith Nair}
\author[1,2]{Alexander Wong}
\affil[1]{DarwinAI, Waterloo, Ontario, Canada}
\affil[2]{University of Waterloo, Waterloo, Ontario, Canada}
\affil[*]{Equal Contribution}

\maketitle

\begin{abstract}
There can be numerous electronic components on a given PCB, making the task of visual inspection to detect defects very time-consuming and prone to error, especially at scale. There has thus been significant interest in automatic PCB component detection, particularly leveraging deep learning. However, deep neural networks typically require high computational resources, possibly limiting their feasibility in real-world use cases in manufacturing, which often involve high-volume and high-throughput detection with constrained edge computing resource availability. As a result of an exploration of efficient deep neural network architectures for this use case, we introduce PCBDet, an attention condenser network design that provides state-of-the-art inference throughput while achieving superior PCB component detection performance compared to other state-of-the-art efficient architecture designs. Experimental results show that PCBDet can achieve up to 2$\times$ inference speed-up on an ARM Cortex A72 processor when compared to an EfficientNet-based design while achieving $\sim$2-4\% higher mAP on the FICS-PCB benchmark dataset.
\end{abstract}

\section{Introduction}

A crucial process in printed circuit board assembly is the visual inspection of electronic components for potential defects. This can help avoid functional failure of devices, user data leakage, or even system control taken by adversaries ~\citet{cryptoeprint:2020/366}. Given that there can be hundreds of electronic components on a given PCB, the task of visual inspection can be extremely time-consuming and prone to operator error, especially during large assembly runs. Therefore, the ability to automatically detect different electronic components on a PCB board for automated inspection purposes is highly desired. As a result, there has been significant interest in the research community in automatic PCB component detection, particularly leveraging deep learning ~\citet{kuo2019data, electronics11081183}. However, one consideration that has been largely left unexplored in research literature in this area is computational efficiency, which is particularly critical for real-world visual quality inspection scenarios involving high-volume, high-throughput electronics manufacturing use-cases under constrained edge computing resources.   

Motivated by the need for both high efficiency and high accuracy for automatic PCB component detection, this study explores efficient deep neural network object detection architectures for the purpose of automatic PCB component detection on the edge.  As a result of this exploration, we introduce PCBDet, a highly efficient, performant self-attention deep neural network architecture design. This architecture notably makes use of the recently introduced AttendNeXt backbone, which has been shown to achieve state-of-the-art performance for TinyML, and is integrated here into RetinaNet~\citet{attendnext, retinanet}.

The paper is organized as follows. In Section 2, details about the architecture of PCBDet, the training procedure, the evaluation procedure, and the data used for training and evaluation are described. In Section 3, experimental results in terms of component detection performance, model size, and inference speed on different computing hardware are presented. Finally, conclusions are drawn and future directions are discussed in Section 4. 

\section{Methods}

\begin{figure}[t]
        \makebox[\textwidth][c]
		{\includegraphics[scale = 0.37]{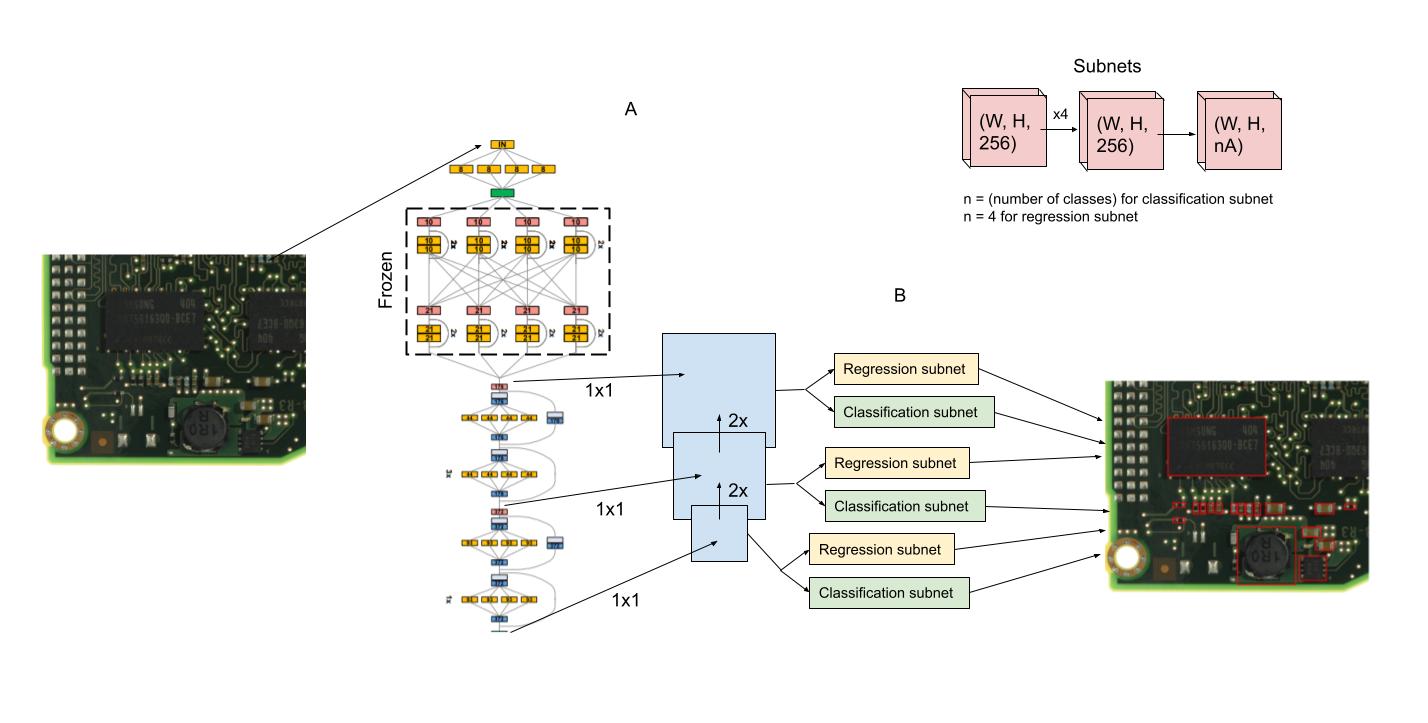}}%
	\caption{Overview of PCBDet architecture. PCBDet consists of A) a double-condensing attention condenser feature encoder feeding a FPN, and B) classification and box regression convolutional sub-nets for bounding box prediction. }
	\label{fig1}
\end{figure}

\subsection{Dataset}

To appropriately explore the quality and impact of our network design, a dataset that can facilitate the training and validation of robust models is essential. With this in mind, models were trained and tested using the DSLR images of the FICS-PCB dataset, a public, comprehensive, and diverse PCB component dataset that includes a number of challenging cases. The dataset itself consists of a total of 31 PCB boards containing over 77 thousand components to detect, with capacitors and resistors being the most widely represented pieces~\citet{cryptoeprint:2020/366}. Each board in the FICS-PCB dataset was further truncated into square patches, reducing train-time resource demand but maintaining component-wise resolution, with each patch being reshaped for additional size reduction further in the input pipeline. Figure 2 demonstrates several examples of such patches with annotated ground truth component labels.  

Division of the dataset into distinct train, validation, and test splits is another crucial element in confirming the soundness of our experimentation. The preservation of exclusivity in the test set here is integral, since it allows for performance evaluation on a strict holdout set, and thus all of the image patches extracted from seven of the 31 PCBs were used as the test set. With the exception of one of the boards, which was excluded due to a lack of DSLR pictures, patches from the remaining boards (which total to 23) were used for the train/validation splits. From this set, 87.5\% of the patches were taken for the train set, while the other 12.5\% were used for the validation set (for post-epoch performance validation).

\begin{figure}[t]
	\makebox[\textwidth][c]
		{\includegraphics[scale = 0.32]{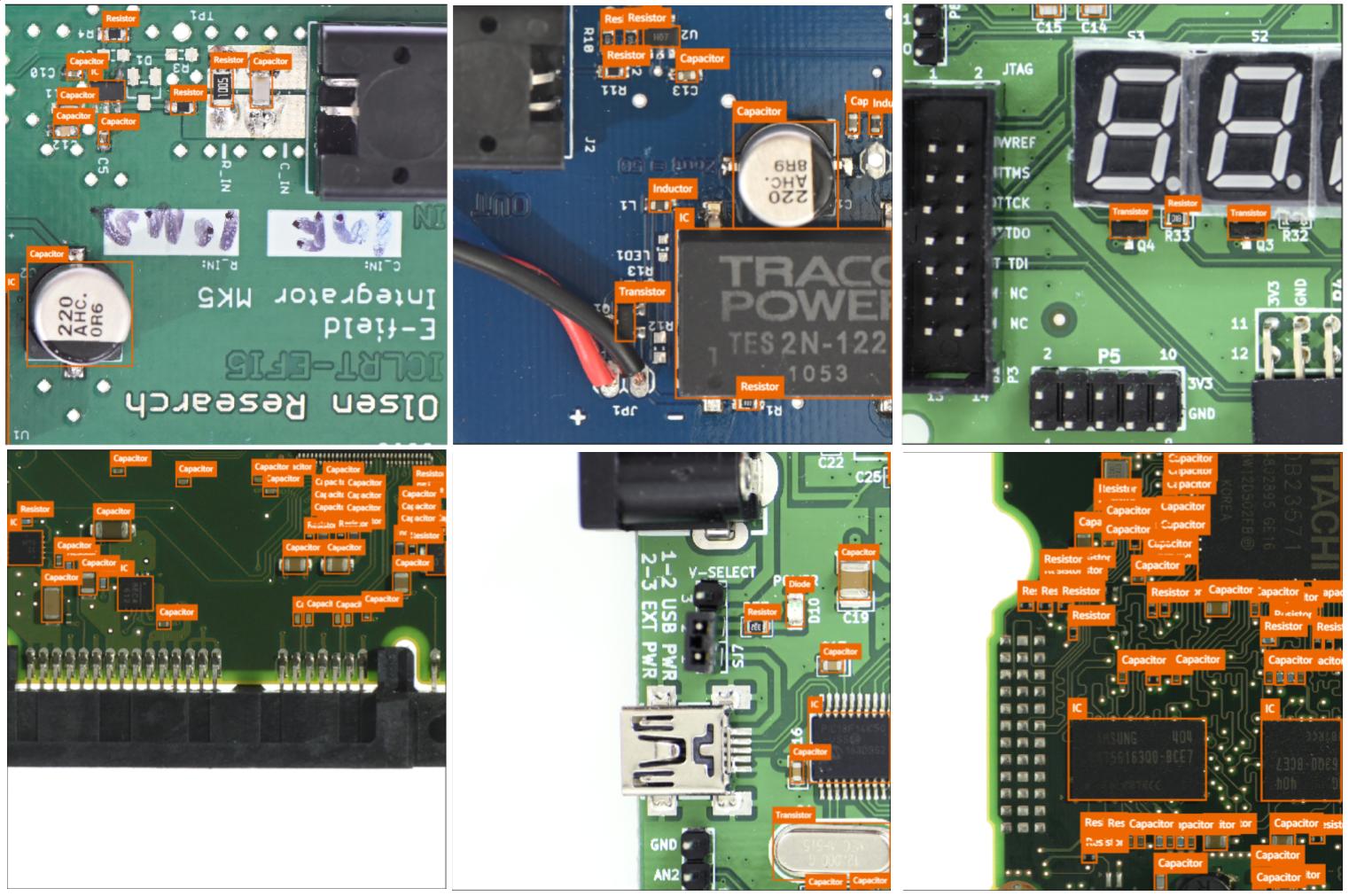}}%
	\caption{Examples of PCB patches annotated with ground truth bounding boxes.}
	\label{fig2}
\end{figure}

\subsection{Architecture}

As seen in Figure 1, the proposed PCBDet possesses an efficient self-attention architecture design inspired by two different architecture design paradigms: 1) RetinaNet bounding box prediction structure~\citet{retinanet}, and 2) double-condensing attention condenser architecture~\citet{attendnext}. As a single-stage detector architecture design, the RetinaNet structure encompasses a more efficient object detection process when compared to state-of-the-art two-stage object detectors like R-CNN~\citet{retinanet}. RetinaNet has also seen increased performance when compared to one-stage detectors such as SSD or YOLO~\citet{retinanet}. As such, the proposed PCBDet architectural design takes inspiration from the RetinaNet structure, looking to adopt an efficient single-stage approach without seeing substantial tradeoffs in performance. 

Without an efficient backbone, however, our network cannot maximize on the possible efficiency-based benefits of the RetinaNet framework. More specifically, the backbone architecture within a RetinaNet structure is the feature encoder that feeds into the convolutional sub-nets, and while a larger, complex backbone may enable increased performance gains, this can lead to substantial losses in efficiency. The design of a small, efficient backbone architecture is thus crucial in creating an effective and efficient object detection network. 

PCBDet's backbone architecture takes inspiration from the AttendNeXt double-condensing attention condenser architecture design, which has shown top of the line performance among other state-of-the-art efficient architectures for ImageNet classification~\citet{attendnext}. This self-attention architecture design features double-condensing attention condenser modules, applied within a convolutional network structure, to increase the speed and efficiency of standard convolutional architectures for feature extraction, thus serving as the basis for an efficient backbone for RetinaNet~\citet{attendnext}. The AttendNeXt feature encoder used in our study was first pretrained on ImageNet, establishing a basis for the weights to be used in our object detection task. The classification head was then removed, and stage outputs were taken as inputs for a feature pyramid network (FPN), whose layers respectively feed into classification and regression subnets as per general RetinaNet structure. To further increase the efficiency of our designed network, the first of four stages of the feature encoder was omitted from the construction of the FPN, decreasing the amount of subnet operations performed per pass. The resultant network from this design process is dubbed PCBDet. 

We also tried integrating an EfficientNetB4-based backbone into RetinaNet and compared the performance of this architecture with PCBDet. EfficientNets are a family of convolutional models generally designed for, as the name implies, efficiency, with the model seeing upscaling as it progresses from B0 through B7~\citet{efficientnet}. EfficientNet-B4 shows improved top-1 ImageNet performance when compared to other state-of-the-art convolutional classifiers while maintaining a lower number of parameters, and was thus chosen as an efficient but potent backbone to explore with RetinaNet~\citet{efficientnet}. As with PCBDet's AttendNeXt backbone, the EfficientNet feature encoder had its classification head removed and block outputs were used as inputs for an FPN. To provide a fair point of comparison for PCBDet, the integration of this feature encoder with the FPN once again seeks to achieve greater efficiency, with the first three of seven blocks of the EfficientNet-B4 feature encoder being excluded from FPN construction. The network designed here is referred to as EfficientNet-Det. 

\subsection{Training}

Proper exploration of our architectures requires thorough training, and for compact networks such as PCBDet in particular, slower, gradual weight learning is crucial to appropriately search for effective weights in the training process.
As such, PCBDet and EfficientNet-Det were each trained for 300 epochs, with a base learning rate of 2e-4 and a proprietary learning rate scheduler, along with Adam optimization. While potential overfitting could arise from slow, gradual learning, this issue was combated with the use of image augmentation, including vertical and horizontal flipping and translation, colour degeneration, and random cutouts (patch removal), as well as the monitoring of network performance on the validation set. It is also essential to address the disproportionate representation of components in the FICS-PCB dataset. To do so, network training uses the focal loss metric, which accounts for class imbalances by adding a focusing parameter to the standard cross-entropy loss, resulting in greatly decreased weighting for easy, well-classified data points~\citet{retinanet}.

During training, the first and second blocks of the AttendNeXt feature encoder in PCBDet were frozen, allowing for the encoder to retain its memories of low-level features from ImageNet pretraining while also tuning higher-level feature blocks to better recognize the shapes and objects associated with PCB components. Similarly, the first four of seven blocks were frozen for the EfficientNet-B4 encoder in EfficientNet-Det. 

\subsection{Evaluation}
Given the goal of efficient model design, we need a method that can effectively measure the complexity and compactness of a model. Inference time, or the time taken per forward pass, is a method that can reveal how quickly a network can perform as a predictor; in our work, inference time was taken for both PCBDet and EfficientNet-Det using an Intel Core i7-10750H processor and a NVIDIA GeForce RTX 1650 Ti, both within a Dell XPS 15 9500 laptop, as well as a Jetson Nano and a 64-bit ARM Cortex A72 processor, altogether providing an image of the on-the-edge inference speeds of the two networks. The number of parameters was also taken for each of the two networks, providing an additional measure for model compactness. 

The predictive power for bounding boxes of our networks is another necessary measure to analyze, in order to compare the model performance of the PCBDet and EfficientNet-Det architectures. This performance assessment was realized using the mean average precision (mAP) for bounding box predictions, over IOU thresholds from 0.5 to 0.95 with a 0.05 step size, commonly known as mAP@[0.5:0.95]; this mAP metric is also known as COCO mAP, the standard performance metric for COCO challenges~\citet{coco}.
Averaging performance over a range of IOU thresholds provides a more generalized sense of object detection performance across resolutions, as lower IOU thresholds test for more roughly correct box predictions while higher thresholds solely reward exact bounding box location. The validation and test performances of PCBDet and EfficientNet-Det were taken to be the mAP@[0.5:0.95] on the validation and holdout test sets respectively; this general AP metric helps to determine the predictive accuracy of our networks during and after training. 

While individual differences in network performance and compactness can be seen through inference time and COCO mAP measures, a collective assessment can provide a better picture of the difference in the accuracy-complexity balance achieved in PCBDet and EfficientNet-Det. This unified analysis can be performed using the NetScore metric, which acts as a quantitative method of assessing this very balance~\citet{netscore}. In this calculation, the coefficient values used were $\alpha$ = 2, $\beta$ = 1, and $\gamma$ = 1, in accordance with the original NetScore study~\citet{netscore}. The inference-time multiply-accumulate (MAC) operations measure was also replaced with the experimental inference time (in seconds) using the ARM Cortex A72 in the calculation, as this experimental metric of complexity provides a more practical measure for edge performance than MAC operations, while the COCO mAP was used as the accuracy metric for the calculation. The calculation used for NetScore was 

\[NetScore = \frac{(mAP*100)^{2}}{(MParams)(Inference \; \, time \;  (s))}\]

\section{Results}
The efficacy of the proposed PCBDet for PCB component detection is compared here with EfficientNet-Det across the following metrics: 1) COCO mAP, 2) model size, and 3) inference speed on various low-power computing hardware. 

\begin{figure}[H]
	\begin{center}
		\includegraphics[scale = 0.8]{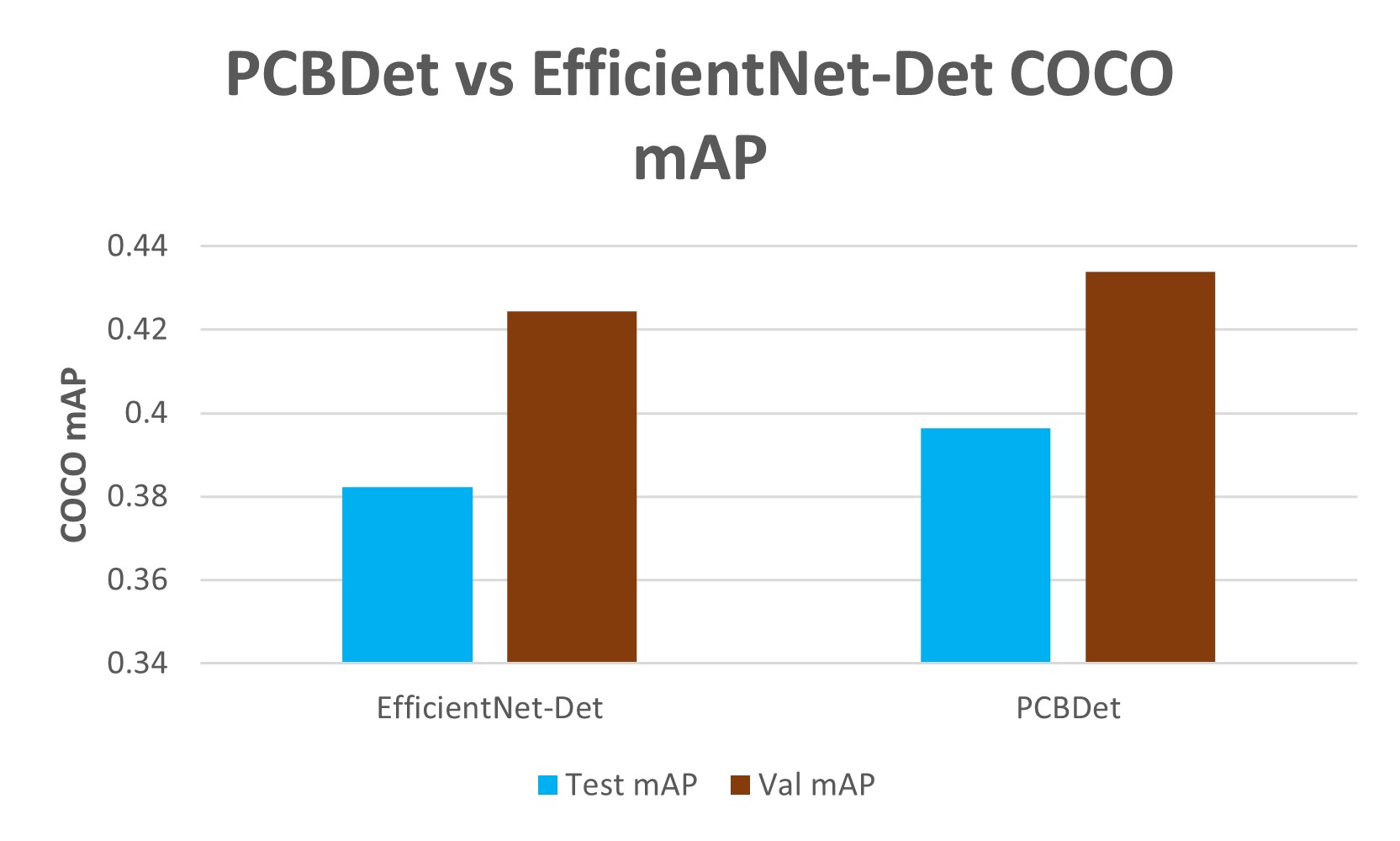}
		
	\end{center}
	\caption{COCO mAP on validation and test data for PCBDet and EfficientNet-Det.}
	\label{fig3}
\end{figure}

\noindent \textbf{COCO mAP}. It can be observed in Figure 3 that the proposed PCBDet achieves noticeable gains in terms of test and validation COCO mAP by approximately 4 \% and 2 \%, respectively,  when compared to EfficientNet-Det.   This gain in mAP is particularly interesting especially given the fact that PCBDet is significantly smaller and faster than EfficientNet-Det, which we will discuss in greater detail. As such, these results illustrate that a high level of performance can be achieved with the proposed PCBDet for the purpose of automatic PCB component detection on the edge.

\noindent \textbf{Model Size}.  As shown in Figure 4, it can be observed that the proposed PCBDet possesses less than half the number of total parameters, as well as trainable parameters, when compared to EfficientNet-Det. This is particularly important for edge based applications such as automatic on-the-edge PCB component detection where memory resources are limited.

\noindent \textbf{Inference Speed}. As shown in Figure 5, it can be observed that the proposed PCBDet is more than 30\% faster than EfficientNet-Det on the NVIDIA Geforce RTX 1650 Ti, with an even greater speed gain on slower hardware such as the Jetson Nano (almost 65\% faster) and Intel Core i7-10750H (over 45\% faster).  As seen in  Figure 6, the performance gains of the proposed PCBDet on lower-power hardware were especially apparent when evaluated on an ARM Cortex A72, where PCBDet was more than 2$\times$ faster than EfficientNet-Det. These inference speed results demonstrate the efficacy of the proposed PCBDet for high-throughput PCB component detection on the edge. 

Finally, using the above results, PCBDet was found to achieve a NetScore of 28.2670 while EfficientNet-Det achieved a NetScore of 13.5749, supporting our findings that PCBDet achieves a superior accuracy-complexity balance when compared to EfficientNet-Det. 

These results demonstrate overall that PCBDet shows significantly greater efficacy than RetinaNet with EfficientNet-B4, which is currently considered a state-of-the-art backbone for TinyML. Ultimately, we have developed a model for PCB object detection that shows very strong performance despite its small size and high inference throughput.  

\begin{figure}[H]
	\begin{center}
		\includegraphics[scale = 0.55]{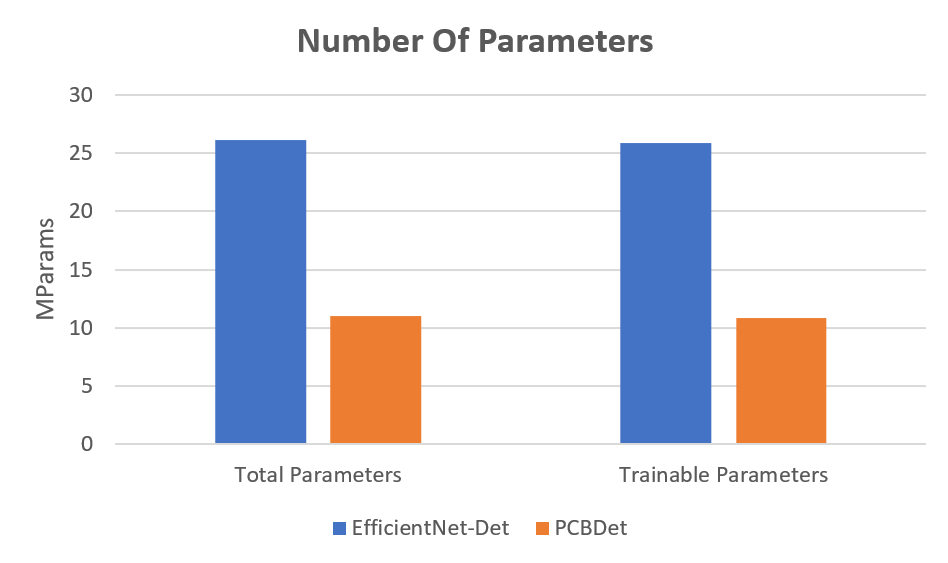}
		
	\end{center}
	\caption{Number of total and trainable parameters for PCBDet and EfficientNet-Det.}
	\label{fig4}
\end{figure}

\bigskip

\begin{figure}[H]
	\begin{center}
		\includegraphics[scale = 0.5]{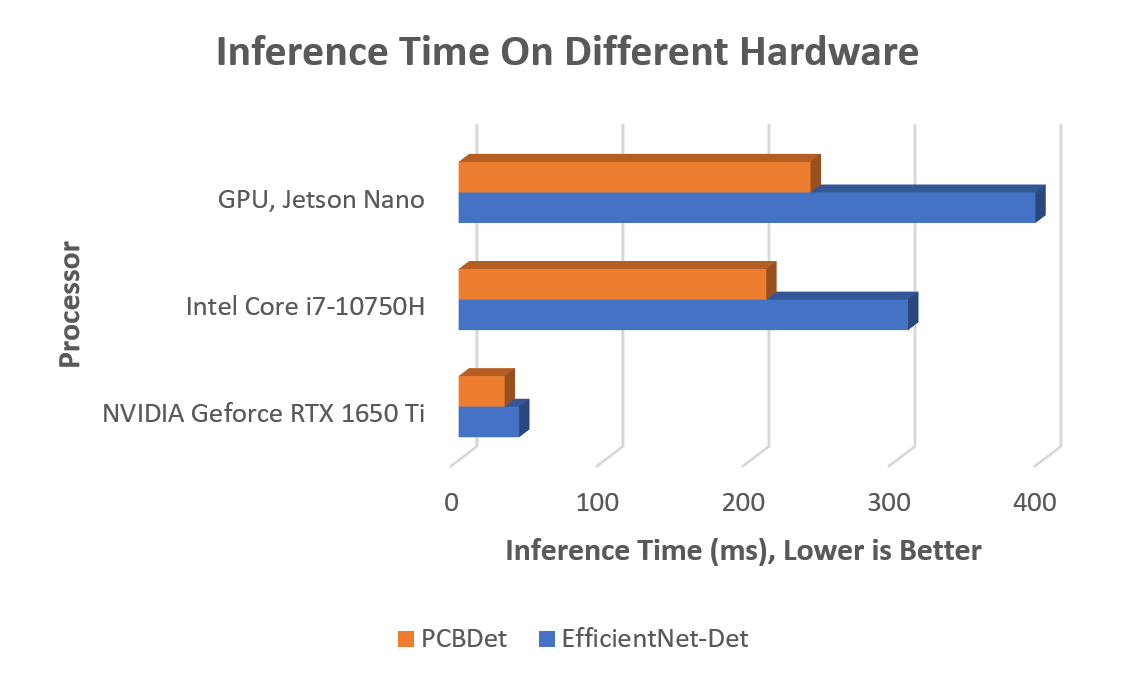}
		
	\end{center}
	\caption{Inference time for PCBDet and EfficientNet-Det across different hardware.}
	\label{fig5}
\end{figure}

\section{Conclusion}
Here, we conducted an exploration of efficient deep neural network object detection architectures for the purpose of automatic PCB component detection on the edge. The resulting network architecture, which we coin PCBDet, methodically integrates the recently introduced AttendNeXt backbone into RetinaNet. This  results in an architecture which can achieve up to a 2x inference speed-up on low power hardware compared to other state-of-the-art efficient architectures, while still achieving a higher mAP on the FICS-PCB benchmark dataset. This makes PCBDet very well-suited for component detection in high-throughput manufacturing scenarios with limited computational resources. Future work may include seeing if a similar strategy involving the methodical use of the AttendNeXt backbone could be employed to develop high performance, efficient deep neural network object detection architectures for other applications. 

\noindent 
\begin{figure}[t]
	\begin{center}
		\includegraphics[scale = 0.4]{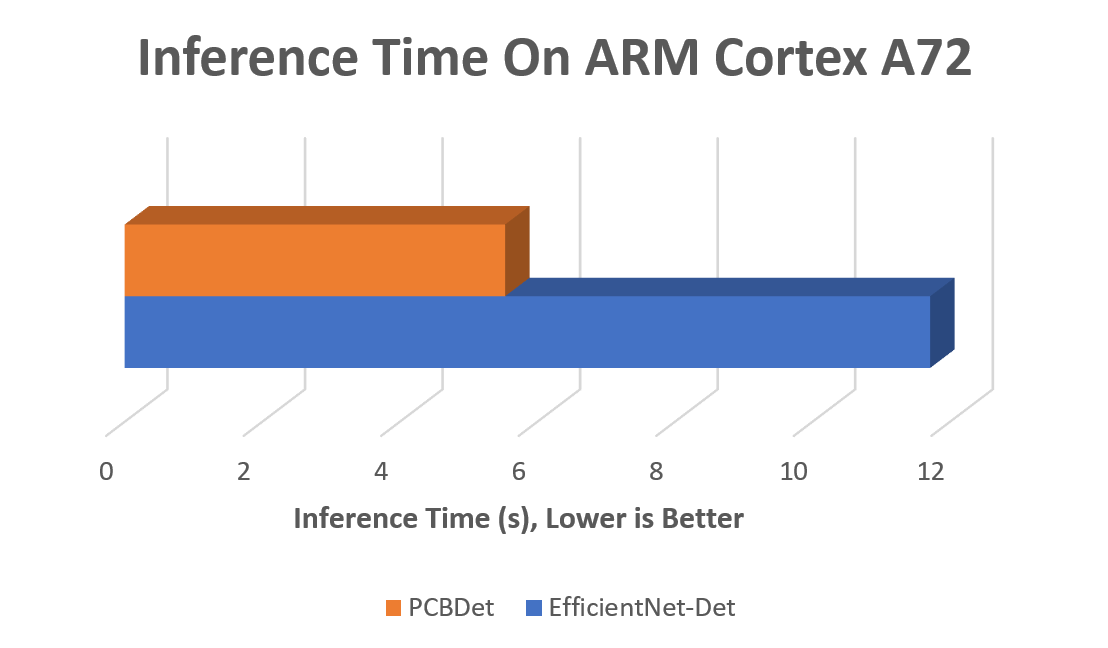}
		
	\end{center}
	\caption{Inference time for PCBDet and EfficientNet-Det on ARM Cortex A72 processor.}
	\label{fig6}
\end{figure}

\bibliographystyle{plainnat}  
\bibliography{references.bib}

\begin{thebibliography}{8}
\providecommand{\natexlab}[1]{#1}
\providecommand{\url}[1]{\texttt{#1}}
\expandafter\ifx\csname urlstyle\endcsname\relax
  \providecommand{\doi}[1]{doi: #1}\else
  \providecommand{\doi}{doi: \begingroup \urlstyle{rm}\Url}\fi

\bibitem[Kuo et~al.(2019)Kuo, Ashmore, Huggins, and Kira]{kuo2019data}
Chia-Wen Kuo, Jacob Ashmore, David Huggins, and Zsolt Kira.
\newblock Data-efficient graph embedding learning for pcb component detection.
\newblock In \emph{2019 IEEE Winter Conference on Applications of Computer
  Vision (WACV)}, 2019.

\bibitem[Li et~al.(2022)Li, Chen, Li, and Gu]{electronics11081183}
Jing Li, Yingqian Chen, Weiye Li, and Jinan Gu.
\newblock Balanced-yolov3: Addressing the imbalance problem of object detection
  in pcb assembly scene.
\newblock \emph{Electronics}, 2022.

\bibitem[Lin et~al.(2017)Lin, Goyal, Girshick, He, and Doll{\'{a}}r]{retinanet}
Tsung{-}Yi Lin, Priya Goyal, Ross~B. Girshick, Kaiming He, and Piotr
  Doll{\'{a}}r.
\newblock Focal loss for dense object detection.
\newblock \emph{CoRR}, 2017.

\bibitem[Lu et~al.(2020)Lu, Mehta, Paradis, Asadizanjani, Tehranipoor, and
  Woodard]{cryptoeprint:2020/366}
Hangwei Lu, Dhwani Mehta, Olivia Paradis, Navid Asadizanjani, Mark Tehranipoor,
  and Damon~L. Woodard.
\newblock Fics-pcb: A multi-modal image dataset for automated printed circuit
  board visual inspection.
\newblock 2020.

\bibitem[Tan and Le(2019)]{efficientnet}
Mingxing Tan and Quoc~V. Le.
\newblock Efficientnet: Rethinking model scaling for convolutional neural
  networks.
\newblock \emph{CoRR}, 2019.

\bibitem[Tong et~al.(2020)Tong, Wu, and Zhou]{coco}
Kang Tong, Yiquan Wu, and Fei Zhou.
\newblock Recent advances in small object detection based on deep learning: A
  review.
\newblock \emph{Image and Vision Computing}, 2020.

\bibitem[Wong(2018)]{netscore}
Alexander Wong.
\newblock Netscore: Towards universal metrics for large-scale performance
  analysis of deep neural networks for practical usage.
\newblock \emph{CoRR}, 2018.

\bibitem[Wong et~al.(2022)Wong, Shafiee, Abbasi, Nair, and Famouri]{attendnext}
Alexander Wong, Mohammad~Javad Shafiee, Saad Abbasi, Saeejith Nair, and Mahmoud
  Famouri.
\newblock Faster attention is what you need: A fast self-attention neural
  network backbone architecture for the edge via double-condensing attention
  condensers.
\newblock 2022.

\end{thebibliography}

\end{document}